\documentclass[runningheads]{llncs}

\usepackage{eccv}

\usepackage{amssymb}
\usepackage{marvosym}
\usepackage{eccvabbrv}

\usepackage{graphicx}
\usepackage{booktabs}

\usepackage[accsupp]{axessibility}  

\usepackage{hyperref}

\usepackage{orcidlink}
\usepackage{color}
\usepackage{multirow}
\usepackage{booktabs}
\usepackage{soul}
\usepackage{colortbl}
\usepackage[ruled,linesnumbered]{algorithm2e}
\definecolor{mygray}{gray}{.9}
\definecolor{mypink}{rgb}{.99,.5,.5}
\definecolor{mycyan}{cmyk}{.3,0,0,0}
\def\ie{\emph{i.e.}}
\def\eg{\emph{e.g.}}

\usepackage{color}
\usepackage{multirow}
\usepackage{booktabs}
\usepackage{wrapfig}
\usepackage{soul}
\usepackage{colortbl}
\usepackage[ruled,linesnumbered]{algorithm2e}
\definecolor{mygray}{gray}{.9}
\definecolor{mypink}{rgb}{.99,.5,.5}
\definecolor{mycyan}{cmyk}{.3,0,0,0}

\usepackage[normalem]{ulem}
\definecolor{headerpink}{HTML}{FDECEC}   
\definecolor{headerblue}{HTML}{E8F2FF}   
\definecolor{rowyellow}{HTML}{FFFBE6}    
\definecolor{posgreen}{HTML}{1A801A}     
\definecolor{negred}{HTML}{D10000}       

\begin{document}

\title{Unfold The World: Factorize 4D Properties in Reinforcing Spatial Reasoning} 

\titlerunning{FactoSR}

\author{Yijun Yang$^{1,2,*}$ \and                                         Shenghe Zheng$^{3,*}$ \and                                                   Wenbo Li$^{2,\dagger}$ \and                                      Jianhui Liu$^{4}$ \and                                                 Haoze Sun$^{2}$ \and            Yanbing Zhang$^{2}$ \and                                                  Jiaxiu Jiang$^{2}$ \and
Lin Song$^{2}$ \and                                                 Haoyang Huang$^{2}$ \and
Nan Duan$^{2}$ \and                                                      Lei Zhu$^{1,\mbox{\Letter}}$}

\authorrunning{Y.~Yang et al.}

\institute{The Hong Kong University of Science and Technology (Guangzhou) \\
\and   
Joy Future Academy \and                              The Hong Kong University of Science and Technology                      \and   The University of Hong Kong }  

\renewcommand{\thefootnote}{}                                          \footnotetext{* Equal Contribution, $\dagger$ Project Lead, \Letter~Corresponding Author}                                           \renewcommand{\thefootnote}{\arabic{footnote}}

\maketitle

\begin{center}
\includegraphics[width=0.95\textwidth]{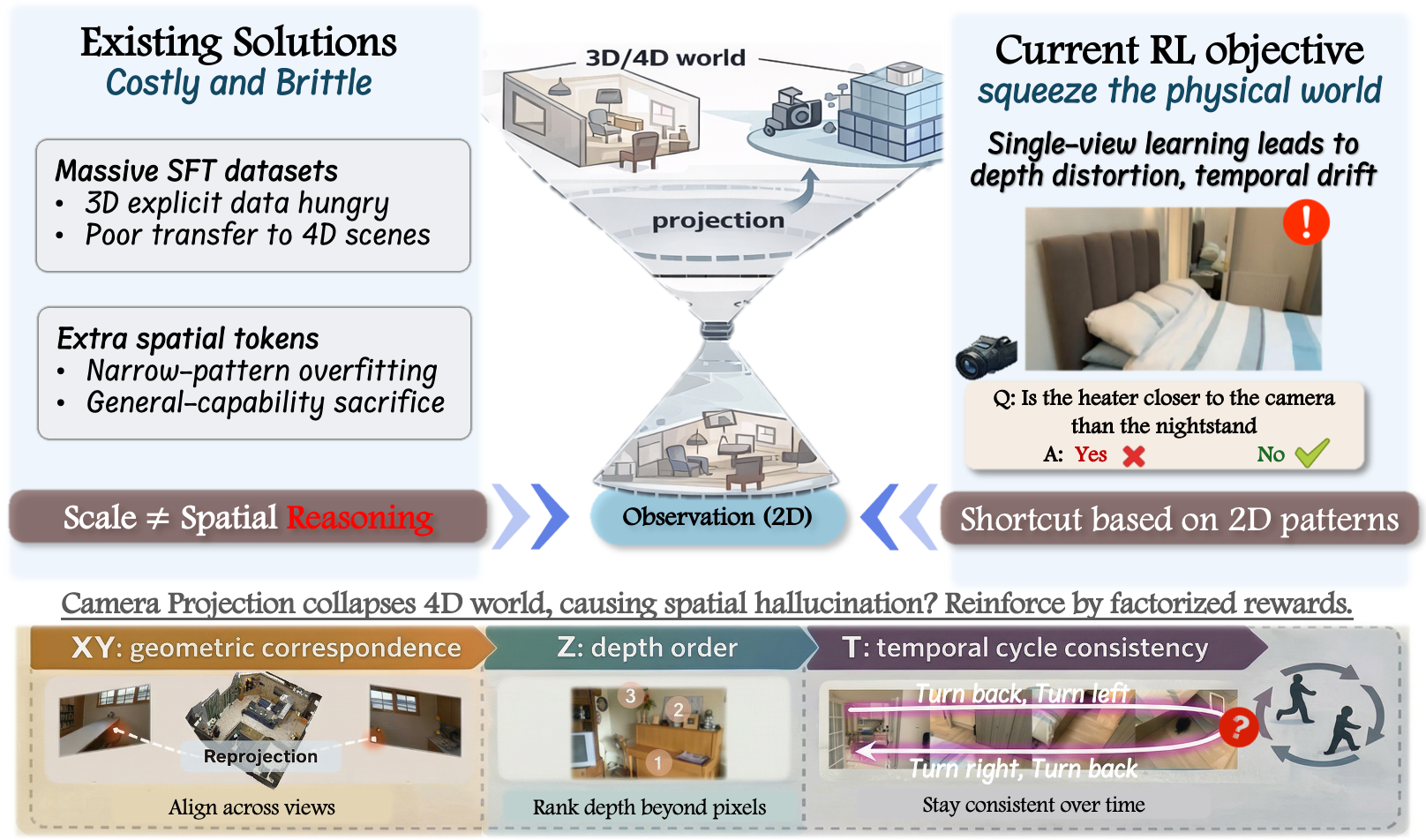}
\vspace{-2mm}
\captionof{figure}{\textbf{Motivation.}
(1) Single-view learning contradicts how the physical world is structured, as real-world perception is inherently multi-view, depth-aware, and temporally continuous.
(2) Scaling SFT or adding spatial tokens does not yield a coherent latent world model, while directly optimizing a 4D objective over space and time is computationally and algorithmically intractable.}
\label{fig:teaser}
\vspace{-4mm}
\end{center}

\begin{abstract}
Despite the remarkable prowess of Vision-Language Models (VLMs) in general multimodal tasks, they remain fundamentally ``flat'' when reasoning about the physical world. 
We argue that this spatial bottleneck stems from a profound dimensional mismatch: while VLMs are trained to interpret 2D projections, true spatial reasoning demands the recovery of latent 3D geometry and temporal continuity. 
To conquer this high-dimensional complexity, we advocate a shift from monolithic learning to a ``divide and conquer'' paradigm. 
We present FactoSR, a factorized reinforcement learning framework that explicitly interpret the dimensions collapsed by visual projection. 
At its core, FactoSR decomposes the monolithic problem of world-consistent reasoning into three orthogonal, geometric sub-objectives: planar correspondence ($XY$), depth consistency ($Z$), and temporal reversibility ($T$). 
By optimizing these verifiable constraints within a unified policy learning mechanism, we effectively transform an ill-posed projection recovery problem into a series of tangible reasoning steps. 
Extensive evaluations on multi-view and video benchmarks demonstrate that this elegant decomposition yields substantial gains in 3D and 4D reasoning, achieving a 5.9\% boost on VSI-Bench and 4.5\% on All-Angles-Bench. 
Our findings suggest that reinforcing explicit, factorized 4D consistency is a critical step toward evolving VLMs into robust, world-aware reasoners.

  \vspace{0.3em} 
  \textbf{Code:} \href{https://github.com/ZimaBlue-WAM/FactoSR}{https://github.com/ZimaBlue-WAM/FactoSR}

  \keywords{Spatial Intelligence \and Reinforcement Learning \and VLMs}
\end{abstract}

\section{Introduction}
\label{sec:intro}

Vision-Language Models (VLMs)~\cite{gpt4, gemini2.5, qwen2vl,internvl2.5, qwen3vl2025, mimovltechnicalreport, ma2026causalspatial} have achieved remarkable success in general visual tasks, yet a critical capability remains elusive, \ie, spatial reasoning. 
This spatial reasoning ability is a fundamental component for VLMs in approaching real-world artificial general intelligence. 
While humans effortlessly identify spatial relationships in sequential visual environments, current VLMs struggle with even basic spatial queries~\cite{cambrian,vsibench,3dsrbench}, which requires interpreting the dynamic world beyond 2D projections. 
This limitation severely constrains their deployment in applications requiring dynamic spatial intelligence, from autonomous driving, robotics navigation to world models.

To mitigate this issue, many studies have synthesized massive spatial question-answering datasets for supervised fine-tuning (SFT)\cite{vst,3dsrbench,cai2025scaling} or explored the incorporation of additional spatial tokens~\cite{vlm3r,wu2025spatialscore,bigverdi2025perception}.
These approaches are often constrained by 3D explicit data hunger, and have poor transferable capability to 4D scenes.
Even more critically, while they sample images from 4D scenes~\cite{scannet,scannet++,arkitscenes,matterport3d}, they heavily rely on single-view-based question answer. 
Their static learning on 2D patterns would induce hallucination in dynamic spatial reasoning, due to the absence of true 4D physical-world perception.
Consequently, they tend to infer spatial relationships heuristically, often making premature decisions without properly accounting for \textit{geometric correspondence, depth estimation, or temporal consistency}, as illustrated in Fig.~\ref{fig:teaser}.

Recently, a few pioneers~\cite{spacer,spatialreasoner,svqa} alternatively investigate applying reinforcement learning with verifiable rewards (RLVR) to enhance the spatial reasoning abilities of VLMs. 
RLVR has demonstrated superior generalization over SFT by learning diverse reasoning strategies rather than static
patterns~\cite{deepseek-r1,vlm-r1,liu2025visual,yang2025medical}.
However, existing RLVR suites for spatial reasoning~\cite{svqa,batra2025spatialthinker,li2025spatialladder,vst,shen2025satori} employ simple rewards inherited from general understanding, which focus only on final correctness. 
They also learn in the single-view spirit, which squeezes the dynamics of the latent world by depth distortion and temporal drift.
Instead, \textit{\uline{spatial intelligence requires explicit reasoning across dimensions of the physical world that are collapsed by the camera projection from reality to observations.}}

Unfortunately, formulating a unified 4D objective over both space and time is computationally and algorithmically intractable.
To divide and conquer~\cite{bertasius2021space}, we present FactoSR, a method beyond pixel-level understanding that factorizes spatial reasoning into plane, depth, and time through online policy reinforcement learning, with three parallel rewards that guide models toward explicit 4D reasoning. 
The training uses a multi-objective reward framework: format rewards ensure structured outputs; accuracy rewards prioritize correctness; \textit{XY} rewards constrain re-projection consistency and correspondences to geometrically valid regions across views; \textit{Z} rewards promote precise 3D localization and relative depth ordering; and \textit{T} rewards enforce temporal cycle consistency and reversible camera-motion reasoning. 
Together with supervised spatial fine-tuning, our suite moves beyond image understanding toward models that reason across the dimensions collapsed by projection, supporting a process of observing, localizing, thinking, and answering.

Building on our constructed datasets, our contributions are threefold.
\begin{itemize}
\item 
\textbf{We present a learning suite that injects spatial knowledge into VLMs.} By cascading supervised fine-tuning (\textit{FactoSR-SFT}) with reinforcement learning (\textit{FactoSR-RL}), we establish a transition from initial spatial perception to explicit reasoning about the latent world.

\item 
\textbf{We introduce a novel factorized reward framework for 4D Reasoning} that decomposes the complicated task into three complementary objectives: \textit{XY-plane}, \textit{Z-depth}, and \textit{T-time}. This design recovers the dimensions typically collapsed by 2D projection, guiding the model to reason precisely across depth and temporal sequences.

\item \textbf{We translate this design into significant spatial gains.}
Our paradigm achieves state-of-the-art performance on multiple spatial benchmarks, particularly an improvement of \textit{5.9\% on VSI-Bench} and \textit{4.5\% on All-Angles-Bench}, while preserving decent general multi-modal capabilities.

\end{itemize}

\section{Related Work}
\label{sec:related}
\noindent\textbf{Spatial Intelligence in MLLMs. }
Recent large vision language models show strong perceptual abilities, yet their spatial intelligence remains limited. Spatial intelligence involves understanding geometric structure, relative positions, and viewpoint transformations, which requires consistent reasoning over spatial configurations rather than surface level recognition. Existing efforts improve spatial intelligence from three aspects. Architectural approaches such as Spatial-MLLM~\cite{spatialmllm}, VLM-3R~\cite{vlm3r}, and 3DThinker~\cite{3dthinker} introduce geometric biases or 3D representations, while SpatialBot~\cite{spatialbot} and VILASR~\cite{VILASR} leverage external perception tools. Data scaling works including SpatialVLM~\cite{spatialvlm}, SpatialRGPT~\cite{spatialrgpt}, VST~\cite{vst}, and SenseNova-SI~\cite{cai2025scaling} expand spatial supervision. Reasoning oriented frameworks such as SpatialLadder~\cite{li2025spatialladder}, Cambrian-S~\cite{yang2025cambrian}, SpaceR~\cite{spacer}, and MindCube~\cite{Mindcube} enhance structured spatial inference. However, these approaches do not provide both depth and temporal analysis of feasible learning pathways for acquiring spatial intelligence. To address this gap, we propose a reinforcement learning framework with decoupled rewards to explicitly strengthen spatial reasoning and guide the model toward more effective behaviors.

\noindent\textbf{Multimodal Reinforcement Learning. }
We focus on enhancing the general reasoning abilities of multimodal models via reinforcement learning. OpenVLThinker~\cite{Openvlthinker} leverages iterative SFT–RL cycles to facilitate reasoning emergence, while VL-Rethinker~\cite{Vl-rethinker} and SRPO~\cite{Srpo} incorporate self-reflection into RL to promote slow thinking. Open Vision Reasoner~\cite{wei2025open} studies cognitive behavior transfer from LLMs, and NoisyGRPO~\cite{qiu2025noisygrpo} and EvolvedGRPO~\cite{shenevolvedgrpo} improve generalization and stability through noise modeling and progressive training. Generative RLHF-V~\cite{rlhf-v} further advances multimodal reward modeling. Overall, prior work moves beyond simple outcome rewards toward more structured and robust RL paradigms for multimodal reasoning.

Reinforcement learning for spatial intelligence is a subfield of multimodal RL, mainly focusing on how to stably improve spatial reasoning through verifiable rewards. SVQA-R1~\cite{svqa} and SpatialThinker~\cite{batra2025spatialthinker} incorporate spatial relations into RL objectives via single-view-consistent or dense spatial rewards. Visionary-R1~\cite{xia2025visionary} and SATORI-R1~\cite{shen2025satori} show that free-form reasoning suffers from shortcut learning, and introduce intermediate verifiable stages such as captioning or region localization. 
Perception-R1~\cite{Perception-r1} highlights the importance of perception-oriented rewards. Meanwhile, Visual Spatial Tuning~\cite{vst} and SpatialLadder~\cite{li2025spatialladder} adopt progressive training from perception to reasoning, while SpatialReasoner~\cite{spatialreasoner} explores explicit 3D representations for better generalization. Overall, prior work suggests that long CoT reasoning or sparse rewards alone is insufficient, motivating our structured solution with explicit representations and multiple verifiable rewards.

\vspace{-2mm}
\section{Preliminaries}
\noindent\textbf{Problem Formulation. }
In this work, we study 4D spatial-temporal intelligence in Vision-Language Models (VLMs), 
which requires joint reasoning over 3D geometric structures and temporal dynamics. Let $D = \{x_1, x_2, \ldots, x_N\}$ denote a spatial-temporal reasoning dataset, where 
each sample $x_i = (V_{1:t}, Q_{st}, a)$ consists of a sequence of visual observations $V_{1:t}$, 
a spatial-temporal query $Q_{st}$, and the corresponding ground-truth answer $a$. Here, $V_{1:t} = \{V_1, V_2, \ldots, V_t\}$ represents a time-ordered visual sequence 
that encodes dynamic 3D scene evolution.

Unlike conventional multimodal reasoning, the 4D intelligence task 
requires constructing an implicit spatio-temporal representation $S(V_{1:t})$, 
which captures geometric attributes (\eg, distance, orientation, occlusion, topology) 
as well as temporal dynamics (\eg, motion, interaction, and state transitions).
Formally, given $x_i \in D$, the VLM aims to generate a textual token sequence $y$ 
by reasoning over $S(V_{1:t})$ to resolve the spatial-temporal query $Q_{st}$, 
where the generated sequence $\hat{y}$ is expected to align with the ground-truth answer $y$.

\noindent\textbf{Reinforcement Learning with Verifiable Rewards. }
RLVR departs from conventional RL by deriving rewards directly from ground-truth correctness rather than relying on a learned reward model. This eliminates the need for auxiliary reward estimation, simplifies the training pipeline, reduces computational cost, and mitigates reward hacking by grounding supervision in objectively verifiable outcomes.

Current implementations of RLVR generally follow a two-part structure: a reward computation scheme that evaluates answer validity, and a policy optimization procedure built upon Group Relative Policy Optimization (GRPO)~\cite{deepseek-r1,deepseekmath}, which performs stable updates through intra-group comparative advantage estimation.
GRPO is a reinforcement learning algorithm derived from PPO that improves policy stability via group-based relative advantage estimation. Its main advantage is that it does not require a value model, reducing memory and computational cost. The training objective of GRPO is to maximize:
\begin{align}
J_{\mathrm{GRPO}}(\theta)
&=
\mathbb{E}_{q \sim P(Q),\, \{o_i\}_{i=1}^{G} \sim \pi_{\theta_{\mathrm{old}}}(O|q)}
\Bigg[
\frac{1}{G}
\sum_{i=1}^{G}
\frac{1}{|o_i|}
\sum_{t=1}^{|o_i|}
\Big(
\min\!\big(
r_{i,t}(\theta) A_i,\,\\
&
\operatorname{clip}\!\big(r_{i,t}(\theta), 1-\varepsilon, 1+\varepsilon\big) A_i
\big)
- \beta D_{\mathrm{KL}}(\pi_\theta \,\|\, \pi_{\mathrm{ref}})
\Big)
\Bigg] \,,
\end{align}

where
\begin{align}
r_{i,t}(\theta)
&=
\frac{
\pi_\theta\!\left(o_{i,t}\mid q,\, o_{i,<t}\right)
}{
\pi_{\theta_{\mathrm{old}}}\!\left(o_{i,t}\mid q,\, o_{i,<t}\right)
} \,, 
A_i
=
\frac{
r_i - \operatorname{mean}\!\left(\{r_1, r_2, \dots, r_G\}\right)
}{
\operatorname{std}\!\left(\{r_1, r_2, \dots, r_G\}\right)
} \,.
\end{align}



Here, $\pi_{\theta}$ and $\pi_{\theta_{\text{old}}}$ denote the updated and previous policies.
$q$ denotes the input query, $\{o_i\}_{i=1}^G$ are $G$ sampled responses, 
$|o_i|$ is the length of the $i$-th response, and $r_i$ is the reward assigned to response $o_i$.

\begin{figure*}[t]
\centering
\includegraphics[width=0.95\textwidth]{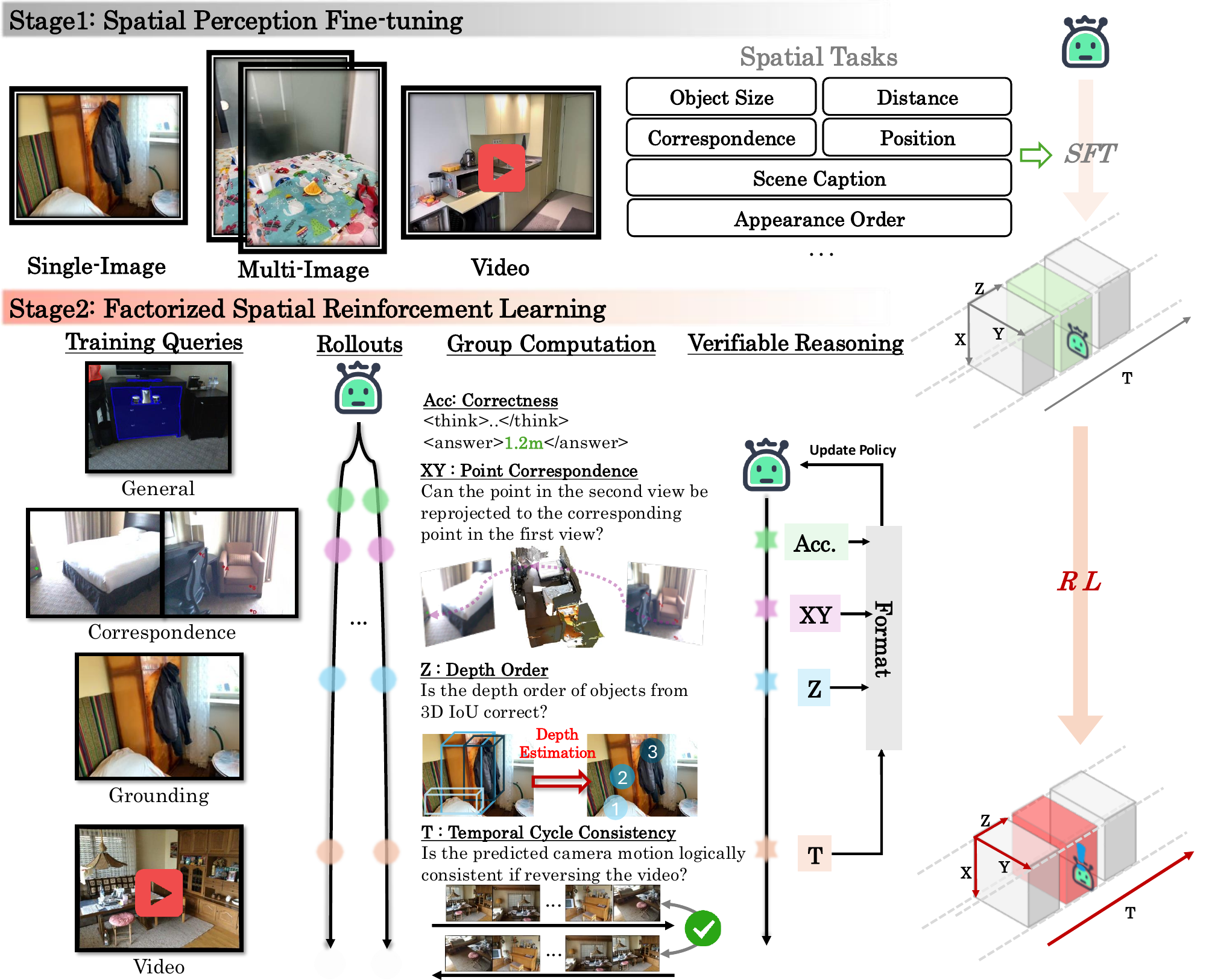}
\vspace{-2mm}
\caption{\textbf{The progressive training framework of FactoSR.} Stage 1 performs supervised fine-tuning on diverse spatial tasks to build foundational spatial perception. Stage 2 introduces factorized reinforcement learning with verifiable rewards (Accuracy, $XY$, $Z$, $T$), explicitly enhancing correspondence, depth, and temporal reasoning.
}
\label{fig:method_main}
\vspace{-6mm}
\end{figure*}

\vspace{-2mm}
\section{Methodology}
\label{sec:method}

\vspace{-2mm}
\subsection{Overview}

Our goal is to endow general vision-language models with explicit spatial reasoning capabilities beyond flat image understanding.
Hence, we build our framework upon a strong general-purpose VLM backbone, Qwen3-VL~\cite{qwen3vl2025}, and train it through a two-stage pipeline, as shown in Fig.~\ref{fig:method_main}.
The data statistics of two stages are introduced in Sec.~\ref{sec:exp_data}.

\noindent\textbf{Stage 1: Spatial Perception Fine-tuning. }
This stage aims to cultivate spatial grounding and foundational perception capabilities through supervised fine-tuning. 
Rather than introducing complex reasoning signals abruptly, we implement a short-to-long supervision curriculum to prepare the model for subsequent RL optimization.

We begin by jointly training the model on short-form paired data, where general multimodal understanding from LLaVA-OneVision~\cite{llava_onevision_data} is synergized with specialized spatial perception tasks. These concise responses emphasize core grounding skills, such as localization, correspondence, and spatial relations, thereby enabling the model to align visual observations with spatial semantics.
During this process, visual tokens extracted from images serve as the conditioning context, and the model is optimized via a standard autoregressive objective:
\vspace{-2mm}
\begin{equation}
\mathcal{L}_{\theta}(y \mid V_{1:t}, Q_{st})
=
-
\sum_{i=2}^{L}
w_i
\log
p_{\theta}
\left(
y_i
\mid
V_{1:t}, Q_{st}, y_{1:i-1}
\right) \,.
\end{equation}
This mixed-task training is performed for a single epoch to inject spatial priors without compromising generalist performance.

Building upon this spatial foundation, we then introduce long-form data featuring structured ``Anchor-Transfer-Verify'' reasoning trajectories for cross-frame alignment, as illustrated in Fig.~\ref{fig:exp_reward_qualitive}. 
The model undergoes further refinement over several hundred iterations using these extended sequences. By progressively scaling from basic grounding to reasoning, Stage I equips the model with incipient spatial logic while ensuring training stability. Crucially, this phase significantly facilitates convergence during the subsequent RL stage.

\noindent\textbf{Stage 2: Factorized Spatial Reinforcement Learning. }
To bridge the gap between supervised awareness and physically consistent reasoning, we employ Group Relative Policy Optimization (GRPO) to further refine the model's reasoning trajectories. Building upon the ``Anchor-Transfer-Verify'' CoT framework from Stage I, this reinforcement stage iteratively optimizes the model's thinking process against our curated verification dataset. 
Specifically, we transition from static supervision to a factorized rule-based reward that evaluates the consistency of the generated reasoning across three physical dimensions: $XY$ (planar correspondence), $Z$ (depth order), and $T$ (temporal cycle-consistency). 
This structured feedback steers the model's latent thinking away from heuristic shortcuts toward a coherent, self-verifying 4D spatial logic.
The detailed formulations of these rewards and the final optimization objective are introduced in the following subsections.

\vspace{-4mm}
\subsection{Factorized Spatial Reinforcement Learning}

In this stage, we employ RL to further enhance the spatial reasoning
capabilities of the stage-2 model. 
For this purpose, we utilize the revised GRPO algorithm~\cite{dapo}, which bypasses the need for a value model by computing the relative advantage of each response within a group of responses to the same question. To facilitate this process, we curated a verification dataset comprising tasks related to spatial understanding, 3D object detection, and general multi-modal understanding. 
In the GRPO framework, we employ a mixed rule-based reward to evaluate the generated responses, including format, accuracy, $XY$, $Z$, and $T$ rewards.
For a given response $\hat{y}$ and its corresponding ground truth $y$, the basic accuracy reward function is used to ensure correctness and is defined as: $\mathcal{R}_{acc}(y,\hat{y}) = \mathbb{I}[\hat{y}=y]$.

\noindent\textbf{XY Reward: Point Correspondence. }
\label{sec:xy_reward}
For multi-view spatial reasoning, a model should not ``guess'' correspondence from 2D appearance alone.
Instead, a correct correspondence must be \emph{geometrically admissible}: the predicted point in the target view should agree with the reprojection of the reference point under camera intrinsics, poses, and depth.
Therefore, our $XY$ reward directly supervises \emph{2D correspondence} by enforcing \emph{reprojection consistency} and \emph{overlap validity}, providing dense, physically grounded guidance during RL.

Given two views $V_1,V_2$ with depth maps $\mathcal{D}_1,\mathcal{D}_2$, intrinsics $\mathbf{K}_1,\mathbf{K}_2$, and camera-to-world poses $\mathbf{T}_1,\mathbf{T}_2$,
we are provided a reference pixel $\mathbf{p}_1=(u_1,v_1)$ in $V_1$ and a discrete candidate set $\{\mathbf{p}_2^{(A)},\mathbf{p}_2^{(B)},\mathbf{p}_2^{(C)},\mathbf{p}_2^{(D)}\}$ in $V_1$.
Let $\hat{\mathbf{p}}_2=(\hat{u}_2,\hat{v}_2)$ be the model-selected candidate point in view 2.
%
Specifically, we first compute the reprojection map from $V_1$ to $V_2$.
For each pixel $\mathbf{p}_1=(u_1,v_1)$ in view 1 with depth $d_1=\mathcal{D}_1(\mathbf{p}_1)$, we unproject to camera coordinates:
\begin{equation}
\mathbf{x}_1 = d_1 \, \mathbf{K}_1^{-1}\,\tilde{\mathbf{p}}_1,
\quad 
\tilde{\mathbf{p}}_1 = (u_1,v_1,1)^\top \,.
\end{equation}
We then transform it to world coordinates and reproject to view 2:
\begin{equation}
\mathbf{X} = \mathbf{T}_1 \, \tilde{\mathbf{x}}_1,
\quad 
\mathbf{x}_2 = \mathbf{T}_2^{-1}\mathbf{X},
\quad
\tilde{\mathbf{p}}_2 \sim \mathbf{K}_2 \mathbf{x}_2 \,,
\end{equation}
where $\tilde{\mathbf{x}}_1=(\mathbf{x}_1^\top,1)^\top$ and $\sim$ denotes equality up to scale, and the resulting pixel coordinate is
$
\mathbf{p}_2^\star = (u_2^\star,v_2^\star)=
\left(
\frac{\tilde{p}_{2,x}}{\tilde{p}_{2,z}},
\frac{\tilde{p}_{2,y}}{\tilde{p}_{2,z}}
\right) \,.
$
We also record the projected depth in view-2 camera coordinates $z_2^\star$ and a validity mask
\begin{equation}
\mathbb{I}_{\mathrm{valid}}(\mathbf{p}_1)=
\mathbb{I}[z_2^\star>0]\cdot 
\mathbb{I}[\mathbf{p}_2^\star\in V_2] \,,
\end{equation}

If $\mathbb{I}_{\mathrm{valid}}(\mathbf{p}_1)=0$, the correspondence is undefined, and we assign a zero reward.
Otherwise, we compare the model prediction $\hat{\mathbf{p}}_2$ with the projected target $\mathbf{p}_2^\star$ in a normalized coordinate system:
\begin{equation}
d \;=\; \left\|
\left(\frac{u_2^\star}{W_2},\frac{v_2^\star}{H_2}\right)
-
\left(\frac{\hat{u}_2}{W_2},\frac{\hat{v}_2}{H_2}\right)
\right\|_2 \,,
\end{equation}
where $(H_2,W_2)$ is the size of $V_2$.
We then assign a soft, distance-aware reward:
\begin{equation}
r_{\mathrm{reproj}} = 
\exp\!\left(-\frac{d}{\sigma}\right)\cdot \mathbb{I}[d \le 3\sigma] \,,
\label{eq:xy_reproj}
\end{equation}
where $\sigma$ controls tolerance.
The hard cutoff $\mathbb{I}[d\le 3\sigma]$ prevents rewarding far-away guesses and stabilizes RL by suppressing spurious gradients from grossly incorrect correspondences.

Reprojection alone may still reward points that are geometrically close but \emph{not visible} in view 2 due to occlusion.
To enforce physical plausibility, we construct an overlap mask on view 2 by checking depth consistency.
For each valid reprojection, we mark $\mathbf{p}_2$ as visible if
\begin{equation}
\left|\mathcal{D}_2(\mathbf{p}_2) - z_2^\star \right| \le \delta \,,
\label{eq:depth_consistency}
\end{equation}
where $\delta$ is a depth threshold.
Thus, we construct the overlap mask $\mathbf{M}_2(\cdot)\in\{0,1\}$, which identifies pixels in view~2 that are truly visible from view~1 under the given camera configuration.
It is derived from the same reprojection process used to compute $\mathbf{p}_2^\star$.
We gate the $XY$ reward using the mask:
\begin{equation}
R_{XY}=
r_{\mathrm{reproj}}\cdot \mathbf{M}_2\!\left(\hat{\mathbf{p}}_2\right) \,.
\label{eq:xy_final}
\end{equation}
This overlap gating turns the reward into a \emph{visibility-aware} signal:
even if $\hat{\mathbf{p}}_2$ is close to $\mathbf{p}_2^\star$, it receives zero reward if it falls outside the depth-consistent overlap region, discouraging correspondences on occluded or non-overlapping areas.
Overall, the $XY$ reward explicitly avoids appearance heuristics and promotes cross-view alignment that is both reprojection-consistent and visibility-valid.

\noindent\textbf{Z Reward: Depth Order. }
\label{sec:z_reward}
After SFT establishes the model’s basic grounding ability, we observe that further optimizing IoU localization during RL yields no benefit for spatial visual question answering.
The core difficulty is not object detection itself, but reasoning about \emph{relative depth} between objects in 3D space from 2D observations.
Rather than supervising metric depth values, we directly optimize the correctness of depth order.

The policy is required to output a set of structured 3D bounding boxes.
All predicted boxes are matched with ground-truth boxes using Hungarian assignment~\cite{kuhn1955hungarian} with a 3D GIoU-based cost. 
For each matched object, we extract the depth of its center in the camera coordinate system, producing the predicted and ground-truth depth sequences.
Given the predicted and ground-truth depth sequences
$\hat{\mathbf{z}}=\{\hat{z}^{(1)},\dots,\hat{z}^{(n)}\}$ 
and 
$\mathbf{z}=\{z^{(1)},\dots,z^{(n)}\}$,
we evaluate whether the predicted front–back relationships between objects are consistent with the ground truth using the Kendall-$\tau$ rank correlation.

Specifically, for every pair of objects $(i,j)$ with $i<j$, we compare the ordering of their depths in the predicted and ground-truth sequences:
\begin{equation}
(\hat{z}^{(i)}-\hat{z}^{(j)})
(z^{(i)}-z^{(j)}) > 0 .
\end{equation}
A pair is considered \emph{concordant} if the predicted and ground-truth orders agree,
and \emph{discordant} if the two orders disagree.
Let $N_c$ and $N_d$ denote the numbers of concordant and discordant pairs, respectively.
The Kendall-$\tau$ coefficient is then defined as:
\begin{equation}
\tau =
\frac{N_c-N_d}{\frac{n(n-1)}{2}} .
\end{equation}

The coefficient $\tau \in [-1,1]$ measures the consistency between predicted and ground-truth depth rankings.
We further normalize it to $[0,1]$ to obtain the depth ordering reward:
$
R_{Z}=
\frac{\tau+1}{2}
$.

Finally, this reward directly reinforce the model to recover the relative depth structure of the scene, which constitutes the core reasoning requirement for 3D spatial understanding.

\noindent\textbf{T Reward: Temporal Cycle Consistency. }
\label{sec:t_reward}
While $XY$ and $Z$ rewards collaboratively reinforce 3D reasoning, they do not guarantee that a model truly understands motion over time.
A model may answer a navigation question using static cues without reasoning about how the camera actually moves.
To explicitly reinforce the temporal dimension, we introduce a $T$ reward that enforces cycle consistency, \ie, reasoning from the start view to the end view must be logically reversible when the process is queried in the opposite direction.

Each training sample, therefore, contains a forward question-answer pair together with a constructed inverse question and its expected answer.
For instance, if the forward solution of camera motion corresponds to ``Turn right, Turn back'', the inverse problem, reasoning from the terminal state back to the start, should yield ``Turn back, Turn left'', ensuring a physically consistent cycle.
During RL, the model rolls out both the forward prediction $\hat{y}$ and the inverse prediction $\hat{y}^{inv}$.
The temporal cycle consistency reward evaluates whether both predictions match the expected answers:
\begin{equation}
\mathcal{R}(y,\hat{y}) = \mathcal{R}_{acc}(y,\hat{y})\cdot\mathcal{R}_{acc}(y^{inv},\hat{y}^{inv}) \,.
\end{equation}
$T$ reward complements the accuracy reward by requiring the model to maintain logical reversibility between forward and inverse motion sequences, fostering a robust understanding of ego-motion and temporal causality.

\noindent\textbf{Final Objective. }
To synergistically integrate the semantic accuracy with granular geometric constraints, we define a factorized total reward function $\mathcal{R}_{\text{total}}$. 
We stipulate that any reinforcement is strictly contingent upon the fulfillment of the format constraint $\mathbb{I}[\mathcal{R}_{\text{format}}=1]$, thereby ensuring the structure of the model output. 
The final objective is formulated as a weighted composition:
\begin{equation}
\mathcal{R}_{\text{total}}(y, \hat{y}) = \mathbb{I}[\mathcal{R}_{format}=1]\cdot \left( \lambda_1 \mathcal{R}_{acc} + \lambda_2 \mathcal{R}_{XY} + \lambda_3 \mathcal{R}_{Z} + \lambda_4 \mathcal{R}_{T} \right) \,,
\end{equation}
where $\lambda_{\{\cdot\}}$ denotes the task-specific importance of correspondence, depth reasoning, and temporal consistency.

By factorizing the reward space into explicit spatial, depth, and temporal dimensions, our RL framework effectively transitions from simple pattern matching to a deeper, physically-grounded understanding of 4D scenes.

\vspace{-2mm}
\section{Experiments}
\label{sec:exp}

\begin{figure*}[t]
\centering
\includegraphics[width=1\linewidth]{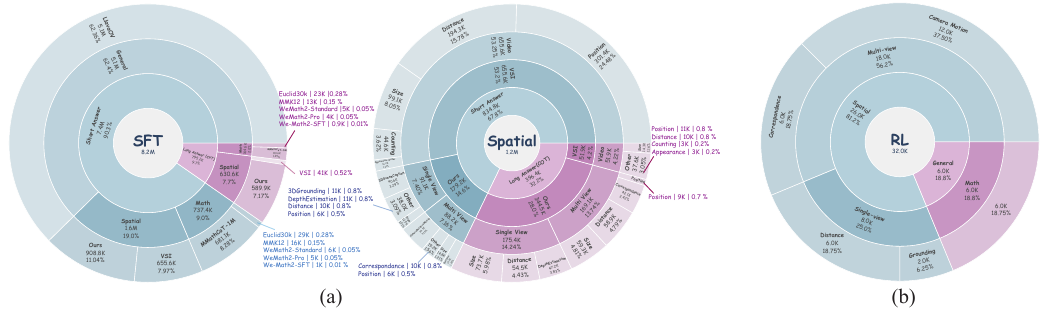}
\vspace{-4mm}
\caption{Overview of (a) FactoSR-SFT and (b) FactoSR-RL datasets. Please zoom in.}
\label{fig:dataset_construction}
\vspace{-6mm}
\end{figure*}

\vspace{-2mm}
\subsection{Data}
\label{sec:exp_data}
\vspace{-2mm}
Fig.~\ref{fig:dataset_construction} summarizes the training data used in our framework.
The nested charts present hierarchical statistics, where each ring from the center outward corresponds to progressively finer-grained categories in the legend.

In the SFT stage, the dataset contains 8.2M samples, dominated by short-answer tasks (7.4M, 90.3\%), while long-answer reasoning data (795K, 9.7\%) provides additional chain-of-thought supervision. The short-answer portion mainly consists of general instruction data LLaVA-OneVision (5.1M)~\cite{llava_onevision_data}, followed by spatial reasoning (1.6M) and mathematical reasoning (737K).
We build a new spatial reasoning dataset using our designed data pipeline. This dataset organizes 1.2M samples across both long-answer and short-answer formats, covering diverse spatial tasks, such as single-view depth estimation, multi-view correspondence, multi-view camera-relative motion, and video spatial understanding~\cite{yang2025cambrian}.

Finally, the RL stage employs a smaller but targeted dataset of 32K samples, where spatial reasoning dominates (81.2\%), complemented by general mathematical reasoning data (18.8\%).

\vspace{-4mm}
\subsection{Implementation Details}
\vspace{-2mm}
\noindent\textbf{Stage 1. }
This training stage aims to establish a strong foundation of spatial understanding capabilities.
For this stage, we use a global batch size of 128, a sequence length of 8,192, and a dynamic data packing strategy to accelerate the training process. 
We employ the AdamW~\cite{adam} optimizer, setting the base learning rate to $5 \times 10^{-5}$ and the vision encoder’s learning rate to $5 \times 10^{-6}$. 
For the CoT cold-start, we continue training the model. The hyper-parameters are adjusted to a global batch size of 128, a base learning rate of $1 \times 10^{-5}$, a vision encoder learning rate of $1 \times 10^{-6}$, and a sequence length of 8,192. 

\noindent\textbf{Stage 2. }
In the RL stage, we refine the model from the first stage using the VeRL~\cite{verl} framework.
We adopt a revised version of the GRPO algorithm~\cite{dapo}, using a rollout size of 8 samples per query and a sampling temperature of 1.0.
This stage utilizes the AdamW optimizer with a constant learning rate of $1 \times 10^{-6}$ and a global batch size of 128.

\noindent\textbf{Evaluation. }
We assess the abilities of state-of-the-art and our models across three distinct capabilities: (1) 4D reasoning ability is benchmarked with All-Angles-Bench~\cite{yeh2025seeing} and VSI-Bench~\cite{vsibench}. We evaluate the fine-grained performance on sub-categories of the two core benchmarks. 
(2) 3D reasoning ability is benchmarked with BLINK~\cite{blink}, 3DSRBench (3DSR\_C)~\cite{3dsrbench}, CVBench (CV-2D, CV-3D)~\cite{cambrian}, ERQA~\cite{ERQA}, RealWorldQA~\cite{grok15v}, and MMSI-Bench (MMSI)~\cite{mmsibench}. 
(3) General multi-modal understanding is evaluated across a suite of standard benchmarks: MMBench~\cite{mmbench} (MMB\_CN, MMB\_EN), MMStar~\cite{mmstar}, and OCRBench~\cite{ocrbench}.
All the models are evaluated using their native system prompt to ensure the fairness of comparisons. More details on prompts, SFT, and RL training setups, are provided in \textit{Appendices}.

\begin{table}[t]
\centering
\caption{Fine-grained comparison on 4D reasoning benchmarks. The best results among open-source VLMs are \colorbox[HTML]{E1F5FE}{\textbf{bold}}. ``Base'' denotes Qwen3-VL-8B-Instruct.}
\vspace{-2mm}
\resizebox{\textwidth}{!}{
\begin{tabular}{l|c|cccccc|c|cccccccccc}
\toprule
\multirow{2}{*}{\textbf{Methods}} 
& \multicolumn{7}{c|}{\textbf{All-Angles-Bench}~\cite{yeh2025seeing}} 
& \multicolumn{11}{c}{\textbf{VSI-Bench}~\cite{vsibench}} \\
\cline{2-19}
& \textbf{Avg.} & Attr. & Pose & Cnt. & Manip. & Rel-Dir. & Rel-Dist.
& \textbf{Avg.} & Cnt. & Obj-Size & Room-Size & Abs-Dist. & Dir-H & Dir-M & Dir-E & Rel-Dist. & Appr-Ord. & Route \\
\midrule
\rowcolor{headerpink} \multicolumn{19}{c}{\textbf{\textit{Open-Source General/Spatial MLLMs}}} \\
\midrule
InternVL3-2B~\cite{internvl3} & 48.6 & 66.3 & 42.6 & 48.2 & 41.8 & 39.5 & 50.2 & 30.4 & 57.7 & 19.8 & 33.3 & 29.0 & 28.7 & 25.7 & 47.5 & 37.0 & 14.7 & 22.7 \\
InternVL3-8B~\cite{internvl3} & 50.5 & 78.6 & 36.4 & 51.0 & 42.0 & 34.7 & 52.8 & 38.7 & 55.4 & 39.7 & 41.1 & 33.6 & 18.5 & 42.6 & 53.0 & 40.8 & 35.1 & 22.7 \\
SpaceR-7B~\cite{spacer} & 49.8 & 71.8 & \cellcolor[HTML]{E1F5FE}\textbf{51.1} & 44.6 & 42.4 & 36.4 & 51.4 & 44.4 & 53.1 & 60.1 & 37.8 & 28.6 & 40.5 & 47.4 & 48.8 & 43.1 & 40.9 & 33.5 \\
MiMo-VL-7B~\cite{mimovltechnicalreport} & 52.9 & 78.3 & 38.1 & 53.4 & 39.9 & 43.8 & 57.1 & 47.8 & 64.9 & 61.4 & 48.9 & 22.4 & 36.2 & 44.2 & 50.2 & 46.5 & 60.2 & 29.9 \\
Qwen2.5-VL-7B-Instruct~\cite{qwen2.5vl} & 50.1 & 74.7 & 50.0 & 51.0 & 39.1 & 37.2 & 50.6 & 36.0 & 42.3 & 46.2 & 40.2 & 22.1 & 29.0 & 40.7 & 50.7 & 36.3 & 28.5 & 30.4 \\
SpatialThinker-7B~\cite{batra2025spatialthinker} & 47.6 & 70.5 & 43.2 & 41.4 & 38.7 & 39.5 & 49.0 & 33.0 & 41.4 & 37.7 & 41.7 & 13.5 & 24.7 & 40.5 & 48.4 & 40.6 & 27.5 & 29.4 \\
VST-7B-SFT~\cite{vst} & 49.5 & 76.5 & 35.8 & 45.8 & 42.0 & 41.8 & 48.2 & 55.3 & 68.0 & 73.2 & 57.6 & 39.8 & 44.8 & 54.0 & 51.6 & 51.5 & 52.8 & 43.3 \\
VST-7B-Thinking~\cite{vst} & 49.0 & 75.2 & 39.2& 47.4 & 42.4 & 36.9 & 48.0 & 52.6 & 61.9 & 73.0 & 58.3 & 33.2 & 40.5 & 46.8 &50.2 & 50.3 & 53.9 & 41.2 \\
Qwen3-VL-8B-Instruct~\cite{qwen3vl2025} & 49.5 & 76.5 & 21.0 & 51.0 & 37.6 & 43.8 & 53.6 & 55.6 & 63.8 & 73.0 & 56.3 & 44.3 & 42.3 & 52.1 & 53.0 & 53.3 & 57.3 & 32.5 \\
Qwen3-VL-8B-Thinking~\cite{qwen3vl2025} & 50.9 & 78.9 & 24.4 & 47.4& 43.7 & 42.9 & 53.0 & 49.8 & 47.0 & 65.6 & 43.9 & 38.8 & 42.3 & 44.4 & 51.6 & 52.5 & 54.7 & 35.1 \\
\midrule
\rowcolor{rowyellow} \multicolumn{19}{c}{\textbf{\textit{Our Method}}} \\
\midrule
\textbf{FactoSR-8B-SFT} & 51.6 & 77.5 & 40.9 & 50.2 & 45.6 & 38.9 & 50.8 & 58.5 & \cellcolor[HTML]{E1F5FE}\textbf{66.7} & 73.7 & 58.2 & 45.8 & 46.4 & 53.7 & 50.2 & 59.3 & 63.1 & 38.2 \\
\textbf{FactoSR-8B-RL}  & \cellcolor[HTML]{E1F5FE}\textbf{55.4} & \cellcolor[HTML]{E1F5FE}\textbf{79.1} & 44.3 & \cellcolor[HTML]{E1F5FE}\textbf{51.0} & \cellcolor[HTML]{E1F5FE}\textbf{45.2} & \cellcolor[HTML]{E1F5FE}\textbf{44.6} & \cellcolor[HTML]{E1F5FE}\textbf{61.1} & \cellcolor[HTML]{E1F5FE}\textbf{61.5} & 66.4 & \cellcolor[HTML]{E1F5FE}\textbf{75.7} & \cellcolor[HTML]{E1F5FE}\textbf{63.7} & \cellcolor[HTML]{E1F5FE}\textbf{46.7} & \cellcolor[HTML]{E1F5FE}\textbf{53.4} & \cellcolor[HTML]{E1F5FE}\textbf{63.7} & \cellcolor[HTML]{E1F5FE}\textbf{61.8} & \cellcolor[HTML]{E1F5FE}\textbf{60.3} & \cellcolor[HTML]{E1F5FE}\textbf{65.7} & \cellcolor[HTML]{E1F5FE}\textbf{46.9} \\
\rowcolor[HTML]{BFDFFF}$\Delta$  (Ours vs. Base) & \textbf{+5.9} & \textbf{+2.6} & \textbf{+23.3} & \textbf{+0.0} & \textbf{+7.6} & \textbf{+0.8} & \textbf{+7.5} & \textbf{+5.9} & \textbf{+2.6} & \textbf{+2.7} & \textbf{+7.4} & \textbf{+2.4} & \textbf{+11.1} & \textbf{+11.6} & \textbf{+8.8} & \textbf{+7.0} & \textbf{+8.4} & \textbf{+14.4} \\

\bottomrule
\end{tabular}
}
\label{tab:finegrain}
\vspace{-4mm}
\end{table}

\begin{table}[t]
\centering
\caption{Quantitative Comparison with state-of-the-art VLMs on 13 3D/4D spatial benchmarks and general benchmarks.}
\vspace{-2mm}
\resizebox{\textwidth}{!}{
\begin{tabular}{l|c|cc|ccccccc|cccc}
\toprule
\multirow{2}{*}{\textbf{Benchmarks}} &
\multirow{2}{*}{\textbf{3/4D-Avg.}}&
\multicolumn{2}{c|}{\textbf{4D Reasoning}} &
\multicolumn{7}{c|}{\textbf{3D Reasoning}} &
\multicolumn{4}{c}{\textbf{GeneralQA}} \\
\cline{3-15}
&&
VSI & AllAngles &
BLINK & 3DSR\_C & CV-2D & CV-3D & ERQA & RealWorldQA & MMSI &
MMB\_CN & MMB\_EN & MMStar & OCRB \\
\midrule

\rowcolor[HTML]{F2F2F2} \multicolumn{15}{c}{\textbf{\textit{Proprietary Models}}} \\ \midrule
Gemini-2.5-Pro~\cite{gemini2.5} & 64.4 &48.4 & 61.3 & 70.6 & 57.6 & 80.4 & 91.3 & 55.8 & 77.3 & 36.9 & 90.2 & 89.2 & 79.1 & 86.6 \\
GPT-4o~\cite{gpt4}  & 57.7 &34.0 & 52.4 & 65.9 & 44.3 & 75.8 & 83.0 & 57.0 & 76.2 & 30.3 & 83.9 & 84.8 & 65.1 & 80.6 \\

\midrule
\rowcolor{headerpink} \multicolumn{15}{c}{\textbf{\textit{Open-Source General/Spatial VLMs}}} \\ \midrule
InternVL3-2B~\cite{internvl3}  & 50.6 &30.4 & 48.6 & 52.8 & 46.4 & 71.9 & 77.3 & 36.2 & 65.5 & 25.9 & 77.1 & 78.3 & 61.5 & 83.7 \\
InternVL3-8B~\cite{internvl3} & 56.2 &38.7 & 50.5 & 55.7 & 52.7 & 80.6 & 86.0 & 40.5 & 70.6 & 30.9 & 81.9 & 82.0 & 68.2 & 88.0 \\
SpaceR-7B~\cite{spacer}  & 53.3 &44.4 & 49.8 & 54.3 & 47.5 & 73.9 & 76.2 & 40.5 & 64.2 & 29.4 & 80.3 & 83.0 & 61.6 & 85.9 \\
MiMo-VL-7B~\cite{mimovltechnicalreport}  & 58.2 &47.8 & 52.9 & 59.7 & 56.1 & 76.9 & 86.9 & 41.0 & \cellcolor[HTML]{E1F5FE}\textbf{73.5} & 29.3 & 80.9 & 80.8 & \cellcolor[HTML]{E1F5FE}\textbf{71.1} & 84.5 \\
Qwen2.5-VL-7B-Instruct~\cite{qwen2.5vl} & 52.8 &36.0 & 50.1 & 55.3 & 49.0 & 75.6 & 73.8 & 41.0 & 68.1 & 26.5 & 82.3 & 82.3 & 70.9 & 87.9 \\
SpatialThinker-7B~\cite{batra2025spatialthinker}  & 53.2 &33.0 & 47.6 & 55.9 & 47.6 & 78.7 & 83.3 & 39.5 & 67.6 & 25.7 & 79.8 & 81.0 & 63.4 & 87.7 \\
VST-7B-SFT~\cite{vst}  & 60.2 &55.3 & 49.5 & 62.1 & 53.3 & 77.9 & 94.8 & 43.8 & 71.5 & 33.3 & 80.4 & 81.1 & 63.1 & 86.3 \\
VST-7B-Thinking~\cite{vst} & 60.5 &52.2 & 49.0 & 62.5 & \cellcolor[HTML]{E1F5FE}\textbf{57.6} & 78.6 & \cellcolor[HTML]{E1F5FE}\textbf{95.5} & 44.5 & 69.3 & \cellcolor[HTML]{E1F5FE}\textbf{34.9} & 76.5 & 77.2 & 63.9 & 86.9 \\
Qwen3-VL-8B-Instruct~\cite{qwen3vl2025} & 59.1 &55.6 & 49.5 & \cellcolor[HTML]{E1F5FE}\textbf{66.1} & 52.8 & 78.6 & 90.8 & 40.1 & 70.7 & 28.1 & 83.3 & 84.2 & 70.1 & \cellcolor[HTML]{E1F5FE}\textbf{90.3} \\
Qwen3-VL-8B-Thinking~\cite{qwen3vl2025} & 59.9 &49.8 & 50.9 & 62.3 & 54.8 & 78.8 & 93.1 & 42.8 & 73.3 & 30.7 & 82.2 & 82.7 & 74.3 & 85.3 \\

\midrule
\rowcolor{rowyellow} \multicolumn{15}{c}{\textbf{\textit{Our Method}}} \\ \midrule
\textbf{FactoSR-8B-SFT} & 60.3 &58.5 & 51.6 & 62.6 & 51.9 &81.1 & 90.7 &45.3 &71.6 &29.7 &83.2 & 85.1 & 67.4 & 87.7 \\
\textbf{FactoSR-8B-RL}  & \cellcolor[HTML]{E1F5FE}\textbf{62.0} &\cellcolor[HTML]{E1F5FE}\textbf{61.5} & \cellcolor[HTML]{E1F5FE}\textbf{55.4} &66.0 & 52.9 &\cellcolor[HTML]{E1F5FE}\textbf{81.3} & 91.7 &\cellcolor[HTML]{E1F5FE}\textbf{47.3} &71.6 &30.7 &\cellcolor[HTML]{E1F5FE}\textbf{84.1} & \cellcolor[HTML]{E1F5FE}\textbf{85.7} & 69.1 & 87.6 \\

\rowcolor[HTML]{BFDFFF}
$\Delta$ (Ours vs. Base) & \textbf{+2.9} & \textbf{+5.9} & \textbf{+5.9} & -0.1 & \textbf{+0.1} & \textbf{+2.7} & \textbf{+0.9} & \textbf{+7.2} & \textbf{+0.9} & \textbf{+2.6} & \textbf{+0.8} & \textbf{+1.5} & -1.0 & -2.7 \\

\bottomrule
\end{tabular}
}
\vspace{-4mm}
\label{tab:main}
\end{table}

\begin{figure*}[t]
\centering
\includegraphics[width=0.95\textwidth]{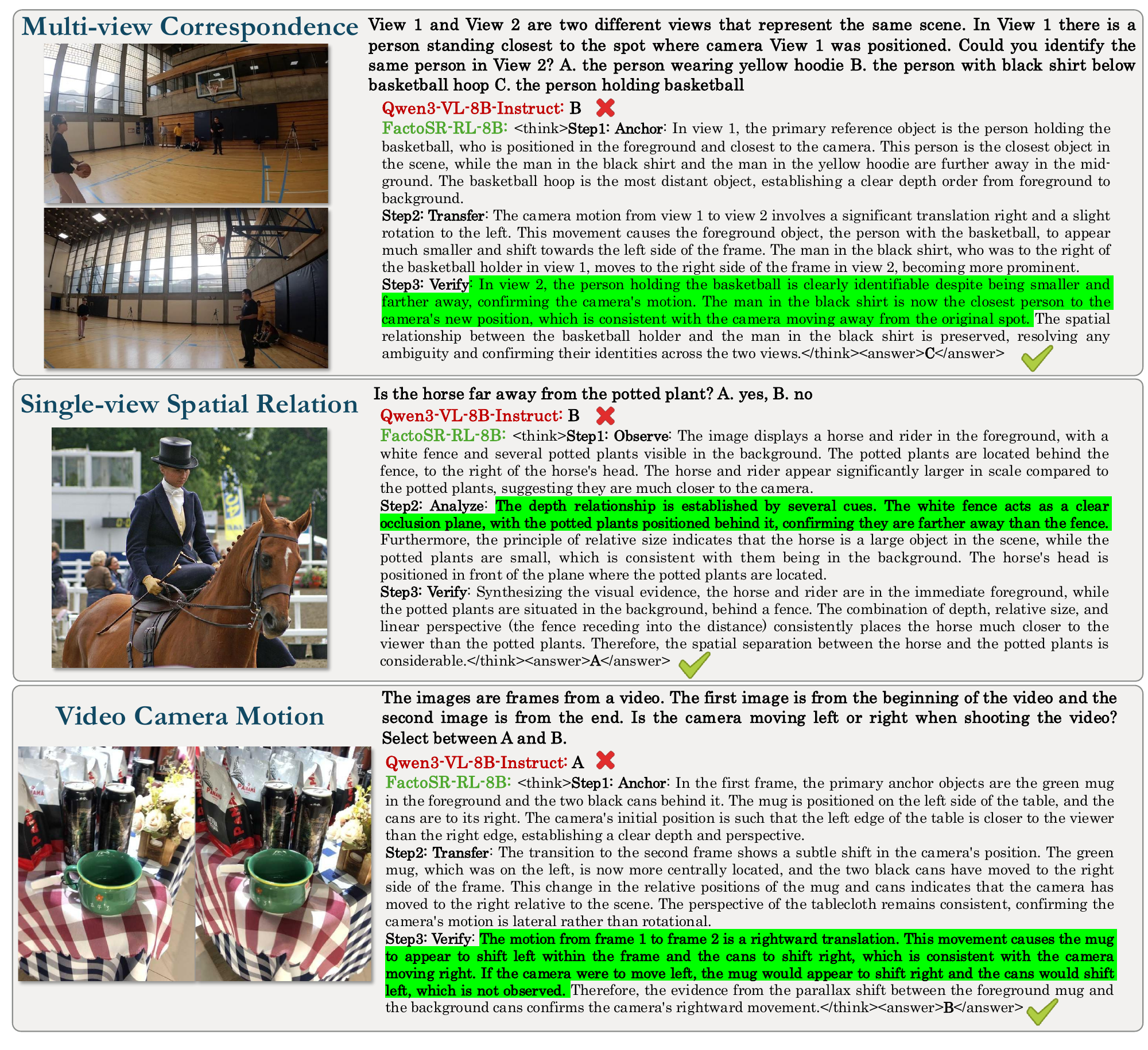}
\vspace{-4mm}
\caption{\textbf{Qualitative comparison between Qwen3-VL-8B Instruct and FactoSR-8B-RL on spatial reasoning tasks.}
While Qwen3-VL-8B Instruct often relies on appearance cues and makes inconsistent judgments, FactoSR-RL produces 4D-consistent reasoning across views, depth, and motion.} 
\label{fig:exp_reward_qualitive}
\vspace{-6mm}
\end{figure*}

\vspace{-2mm}
\subsection{Quantitative Results}
\vspace{-2mm}
\noindent\textbf{4D Reasoning Analysis. }
We compare FactoSR with a range of open-source spatial and general VLMs, including InternVL, Qwen-VL, MiMo-VL, SpaceR, SpatialThinker, and VST. 
As shown in Tab.~\ref{tab:finegrain}, our method achieves the best performance on both All-Angles-Bench and VSI-Bench, two challenging 4D spatial reasoning benchmarks. 
Specifically, FactoSR-8B-RL achieves \textit{61.5\%} on VSI and \textit{55.4\%} on AllAngles, outperforming the strongest open-source baselines by \textit{+5.9} and \textit{+2.5} points, respectively. 
The improvements are consistent across diverse sub-tasks such as attribute identification, camera pose estimation, and relative spatial reasoning on AllAngles, as well as video-based tasks including relative direction, spatial distance reasoning, and route planning on VSI.

\noindent\textbf{Overall Benchmark Comparison. }
Tab.~\ref{tab:main} further compares our model with state-of-the-art proprietary and open-source VLMs across \textit{13} benchmarks.

FactoSR-8B-RL achieves the best spatial reasoning performance with the average of \textit{62.0} across all 3D/4D spatial benchmarks, improving over the base model by \textit{+2.9} points. 
Besides the strong gains on the 3/4D benchmarks, FactoSR also achieves competitive performance on general multimodal benchmarks, \eg, MMBench-CN, MMBench-EN. \textit{More results are provided in Appendices}.

\vspace{-2mm}
\subsection{Ablation Study}
To quantitatively evaluate the contributions of the factorized rewards, we construct three specialized evaluation groups to analyze spatial reasoning improvements: The correspondence group $\mathbf{\Delta_{\text{Cor}}}$ includes All-Angles-Bench (Attribute Identification), BLINK (Functional Correspondence, Semantic Correspondence, Visual Correspondence).
The depth group$\mathbf{\Delta_{\text{Depth}}}$ includes CV-Bench-3D (Depth), BLINK (Relative Depth, Spatial Relation).
The camera motion group $\mathbf{\Delta_{\text{t}}}$ includes RealWorldQA, MMSIBench (Motion-Cam), BLINK (Multi-view Reasoning), VSI-Bench (Route Planning).
The 3D/4D Avg. Acc. is computed over 9 spatial benchmarks.

\begin{wraptable}{r}{0.6\textwidth}
\vspace{-5mm}
\centering
\caption{Average accuracy across all 9 benchmarks and performance on three specialized tasks with relative improvements. \textit{FactoSR} models consistently outperform SFT and vanilla GRPO.}
\small
\resizebox{\linewidth}{!}{
\begin{tabular}{l|c|ccc}
\toprule
\textbf{Model} & \textbf{3D/4D Avg. Acc. (9)} & $\mathbf{\Delta_{\text{Cor}}}$ & $\mathbf{\Delta_{\text{Depth}}}$ & $\mathbf{\Delta_{\text{t}}}$ \\ 
\midrule

\rowcolor{headerpink} \multicolumn{5}{c}{\textit{Supervised Fine-Tuning}} \\ 
FactoSR-SFT & 60.3 & 69.9 & 86.3 & 45.7 \\

\rowcolor{headerblue} \multicolumn{5}{c}{\textit{Reinforcement Learning}} \\
+ Vanilla GRPO 
& 60.5$_{\pm 0.1}$ \textcolor{posgreen}{(+0.2)} 
& 71.6$_{\pm 0.8}$ \textcolor{posgreen}{(+1.7)} 
& 85.8$_{\pm 1.1}$ \textcolor{negred}{(-0.5)} 
& 47.1$_{\pm 2.2}$ \textcolor{posgreen}{(+1.4)} \\

+ XY Reward 
& 60.7$_{\pm 0.1}$ \textcolor{posgreen}{(+0.4)} 
& \textbf{72.6}$_{\pm 0.7}$ \textcolor{posgreen}{\textbf{(+2.7)}} 
& 86.6$_{\pm 1.0}$ \textcolor{posgreen}{(+0.3)} 
& 47.2$_{\pm 2.8}$ \textcolor{posgreen}{(+1.5)} \\

+ Z Reward 
& 60.6$_{\pm 0.2}$ \textcolor{posgreen}{(+0.3)} 
& 69.8$_{\pm 0.6}$ \textcolor{negred}{(-0.1)} 
& \textbf{87.6}$_{\pm 0.4}$ \textcolor{posgreen}{\textbf{(+1.3)}} 
& 47.4$_{\pm 2.0}$ \textcolor{posgreen}{(+1.7)} \\

+ T Reward 
& 60.9$_{\pm 0.2}$ \textcolor{posgreen}{(+0.6)} 
& 69.9$_{\pm 0.3}$ \textcolor{posgreen}{(+0.0)} 
& 87.4$_{\pm 0.8}$ \textcolor{posgreen}{(+1.1)} 
& \textbf{53.6}$_{\pm 1.1}$ \textcolor{posgreen}{\textbf{(+7.9)}} \\ 
\midrule

\rowcolor{rowyellow}
+ Full Rewards 
& \textbf{62.0}$_{\pm 0.2}$ \textcolor{posgreen}{\textbf{(+1.7)}} 
& 72.1$_{\pm 0.7}$ \textcolor{posgreen}{(+2.2)} 
& 87.1$_{\pm 0.6}$ \textcolor{posgreen}{(+0.8)} 
& 50.9$_{\pm 1.3}$ \textcolor{posgreen}{(+5.2)} \\

\bottomrule
\end{tabular}
}
\vspace{-10pt}
\label{tab:main_ablation}
\end{wraptable}

Tab.~\ref{tab:main_ablation} shows that vanilla GRPO yields only marginal gains (+0.2), indicating that general RL signals alone are insufficient for spatial reasoning.
Introducing factorized rewards leads to targeted improvements: XY mainly enhances correspondence reasoning ($\Delta_{\text{Cor}}$ +2.7), Z significantly improves depth understanding ($\Delta_{\text{Depth}}$ +1.3), and T notably strengthens temporal reasoning ($\Delta_t$ +7.9).
Further analysis in Tab.~\ref{tab:ablate_z} reveals that the vanilla grounding reward~\cite{vst} is ineffective, causing overall performance drops (-0.2) and degraded spatial relation reasoning (-0.7).
In contrast, Z reward consistently improves all depth-related metrics, especially Relative Depth (+3.2), validating that RL should emphasize relative depth of objects rather than the simple localization.
Combining all rewards achieves the best overall improvement (1.7\%), demonstrating that XY, Z, and T provide complementary supervision.

\vspace{-6mm}
\begin{table}[h]
\centering
\caption{Ablation on depth supervision showing the effectiveness of the Z reward and vanilla grounding reward over SFT models.}
\vspace{-4mm}
\small 
\renewcommand{\arraystretch}{1.1} 
\resizebox{0.9\linewidth}{!}{
\begin{tabular}{l|c|c|ccc}
\toprule
\textbf{Model} & \textbf{3D/4D Avg. Acc. (9)} &$\mathbf{\Delta_{\text{Depth}}}$ & Depth & Relative\_Depth & Spatial\_Relation \\ \midrule

+ Vanilla Grounding & \textcolor{negred}{-0.2} & \textcolor{negred}{-0.1} & \textcolor{negred}{-0.5} & \textcolor{posgreen}{+0.8} &\textcolor{negred}{-0.7}\\

+ Z Reward & \textcolor{posgreen}{\textbf{+0.3}} & \textcolor{posgreen}{\textbf{+1.3}} & \textcolor{posgreen}{\textbf{+0.2}} & \textcolor{posgreen}{\textbf{+3.2}}& \textcolor{posgreen}{\textbf{+0.6}}\\

\bottomrule
\end{tabular}
}
\vspace{-6mm}
\label{tab:ablate_z}
\end{table}

\vspace{-2mm}
\subsection{Visualization}

\vspace{-2mm}
\noindent\textbf{Analysis of Factorized Rewards. }
Fig.~\ref{fig:exp_reward_qualitive} illustrates how factorized rewards improve structured spatial reasoning.
In multi-view correspondence, SFT models rely on size heuristics and fail to preserve identity. 
After RL, the model follows a cross-view CoT (anchor$\rightarrow$transfer$\rightarrow$verify), correctly tracking the target person across viewpoints, driven by the XY reward.
In single-view spatial relation, the model adopts a depth-aware CoT, reasoning via occlusion and layering rather than size of the potted plant, enabled by the Z reward.
In video camera motion, the model develops a temporal CoT by analyzing coherent foreground–background shifts around the mug, guided by the T reward.
Overall, the factorized rewards promote consistent single-view and multi-view reasoning across space and time.

\begin{figure*}[t]
\centering
\includegraphics[width=0.92\textwidth]{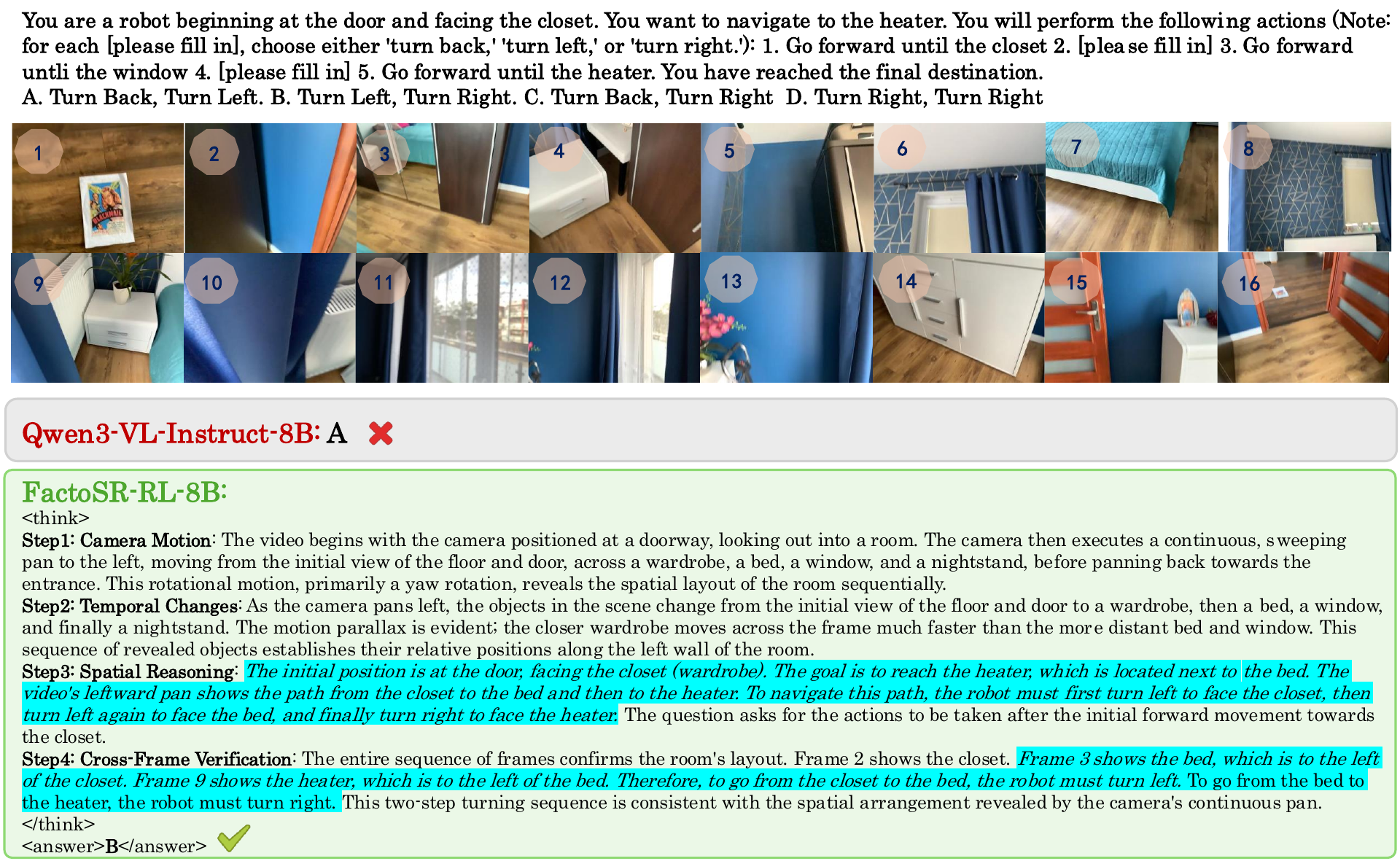}
\vspace{-2mm}
\caption{\textbf{The example of video route plan.} Given a sequence of egocentric observations, our model infers a consistent navigation plan by reconstructing the underlying spatial layout and camera motion, effectively acting as a latent world model.
}
\label{fig:exp_route_plan}
\vspace{-6mm}
\end{figure*}

\noindent\textbf{Generalization to Video Route Plan. }
The visual evidence from the FactoSR-RL-8B model in Fig.~\ref{fig:exp_route_plan} demonstrates that the integration of camera motion data as a structural prior, coupled with a temporal cycle consistency reward in RL framework, significantly generalizes well to video route planning. 
While standard VLMs like Qwen3-VL often fail to maintain spatial orientation, as seen in its false choice of ``Turn Back'', the RL-tuned agent leverages motion parallax and yaw rotation (detailed in Steps 1 and 2) to construct a coherent topological map of the environment. 
Crucially, the model is encouraged to maintain a bijective mapping between temporal video observations and the robot's latent spatial state. 
This ensures that the transition from the closet to the bed, and ultimately to the heater, is grounded in cross-frame verification rather than isolated object recognition. 
As a result, the agent effectively decodes the underlying scene dynamics (Step 4), correctly identifying that ``Turn Left, Turn Right'' is the only path consistent with the reconstructed 4D navigation manifold.

\vspace{-4mm}
\section{Conclusion}
\label{sec:con}
\vspace{-2mm}
Spatial intelligence requires the ability to reason freely within the vast 4D world, yet most vision-language models are shaped under a single-view paradigm that collapses this rich structure into flat image observations. 
To move beyond this limitation, we introduce FactoSR, a reinforcement learning framework that factorizes spatial reasoning into complementary dimensions of plane, depth, and time, allowing models to reason over the latent spatial structure underlying visual observations.
Results on diverse 3D and 4D spatial benchmarks show that FactoSR pushes forward spatial reasoning, encourage multi-modal models to observe, reason, and interact with the latent structure of the physical world.


\section*{Acknowledgements}
This work is supported by the Program of Guangdong Education Department. 
This work is also supported by Joy Future Academy and JD.com for research funding and computing resources.

%
%
\bibliographystyle{splncs04}
\bibliography{main}
\end{document}